\documentclass[10pt,twocolumn,letterpaper]{article}
\usepackage{mmstyle}

\usepackage[pagenumbers]{cvpr} 
\usepackage[table]{xcolor}
\usepackage{indentfirst}

\usepackage{epigraph} 
\usepackage{blindtext}
\usepackage{booktabs}

\def\zz{\mathbf{z}}

\def\mM{\mathcal{M}}

\def\tT{\mathcal{T}}

\def\Re{\mathbb{R}}

\DeclareMathOperator*{\argmin}{arg\,min}

\newcommand\paren[1]{\left(#1\right)}

\def\latex/{\LaTeX}
\def\bibtex/{\hologo{BibTeX}}

\newcommand{\RN}[1]{%
  \textup{\uppercase\expandafter{\romannumeral#1}}%
}

\usepackage{multirow}

\newcommand{\rankone}{add8e6}
\newcommand{\ranktwo}{b5dce9}
\newcommand{\rankthree}{c6e4ee}

\definecolor{cvprblue}{rgb}{0.21,0.49,0.74}
\definecolor{cvprred}{rgb}{0.74,0.21,0.49}
\usepackage[pagebackref,breaklinks,colorlinks,citecolor=cvprblue,linkcolor=cvprred]{hyperref}

\def\paperID{7297} 
\def\confName{CVPR}
\def\confYear{2024}

\title{GPT-4V(ision) is a Human-Aligned Evaluator for Text-to-3D Generation}

\author{
Tong Wu$^{1,5}$\footnotemark[1]
\and
Guandao Yang$^{2}$\footnotemark[1]
\and
Zhibing Li$^{1,5}$\footnotemark[1]
\and
Kai Zhang~$^{3}$
\and  
Ziwei Liu~$^{4}$
\and
Leonidas Guibas~$^{2}$
\and 
Dahua Lin~$^{1,5}$
\and 
Gordon Wetzstein~$^{2}$
\and
{\small
$^{1}$ The Chinese University of Hong Kong \quad
$^{2}$ Stanford University \quad
$^{3}$ Adobe Research\quad} \\
{\small 
$^{4}$ S-Lab, Nanyang Technological University\quad
$^{5}$ Shanghai Artificial Intelligence Laboratory
}
}

\begin{document}
\twocolumn[{%
\renewcommand\twocolumn[1][]{#1}%
\maketitle
\begin{center}
    \centering
    \captionsetup{type=figure}     
    \vspace{-1em}
    \includegraphics[width=\textwidth]{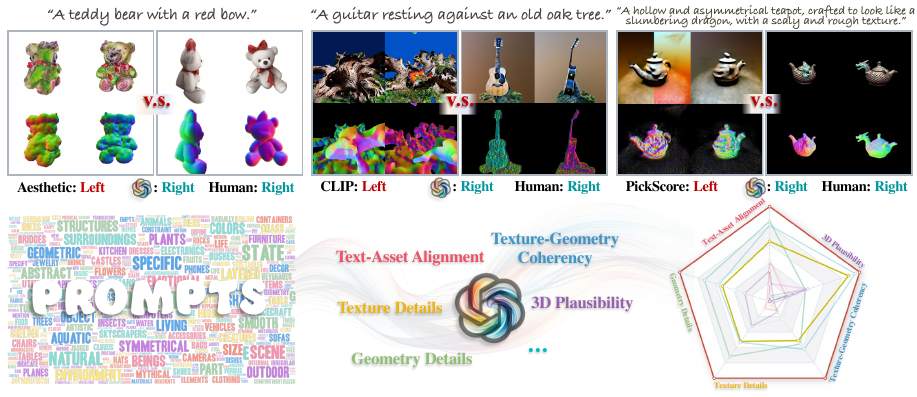}
     \vspace{-2em}
    \captionof{figure}{
    We present a versatile and human-aligned evaluation metric for text-to-3D generative methods. 
    To this end, we design a prompt generator that can produce a set of input prompts targeting an evaluator's demands. 
    Moreover, we leverage GPT-4V to compare two 3D shapes according to different evaluation criteria.
    Our method provides a scalable and holistic way to evaluate text-to-3D models.
    }
    \label{fig:teaser}
\end{center}%
}]


\begin{abstract}
\vspace{-1em}
\let\thefootnote\relax\footnotetext{\noindent{$^*$ Equal contribution.}}
Despite recent advances in text-to-3D generative methods, there is a notable absence of reliable evaluation metrics.
Existing metrics usually focus on a single criterion each, such as how well the asset aligned with the input text.
These metrics lack the flexibility to generalize to different evaluation criteria and might not align well with human preferences.
Conducting user preference studies is an alternative that offers both adaptability and human-aligned results.
User studies, however, can be very expensive to scale.
This paper presents an automatic, versatile, and human-aligned evaluation metric for text-to-3D generative models.
To this end, we first develop a prompt generator using GPT-4V to generate evaluating prompts, which serve as input to compare text-to-3D models.
We further design a method instructing GPT-4V to compare two 3D assets according to user-defined criteria.
Finally, we use these pairwise comparison results to assign these models Elo ratings.
Experimental results suggest our metric strongly aligns with human preference across different evaluation criteria.
Our code is available at 
\href{https://github.com/3DTopia/GPTEval3D}
{https://github.com/3DTopia/GPTEval3D}.
\end{abstract}    
\section{Introduction}
\label{sec:intro}
The field of text-to-3D generative methods has seen remarkable progress over the past year, driven by a series of breakthroughs. 
These include advancements in neural 3D representations \cite{park2019deepsdf, mildenhall2021nerf}, the development of extensive datasets \cite{chang2015shapenet, objaverse, objaverseXL}, the emergence of scalable generative models \cite{ho2020denoising, song2020score, rombach2022high}, and the innovative application of text--image foundational models for 3D generation \cite{poole2022dreamfusion,Po:2023:star_diffusion_models}. 
Given this momentum, it's reasonable to anticipate rapidly increasing research efforts and advancements within the realm of text-to-3D generative models.

Despite recent advances, the development of adequate evaluation metrics for text-to-3D generative models has not kept pace. 
This deficiency can hinder progress in further improving these generative models.
Existing metrics often focus on a single criterion, lacking the versatility for diverse 3D evaluation requirements. 
For instance, CLIP-based metrics~\cite{poole2022dreamfusion,jain2022zero} are designed to measure how well a 3D asset aligns with its input text, but they may not be able to adequately assess geometric and texture detail. 
This lack of flexibility leads to misalignment with human judgment in evaluation criteria the metric is not designed for.
Consequently, many researchers rely on user studies for accurate and comprehensive assessment. 
Although user studies are adaptable and can accurately mirror human judgment, they can be costly, difficult to scale, and time-consuming.
As a result, most user studies have been conducted on a very limited set of text-prompt inputs.
This leads to a question: 
\textit{Can we create automatic metrics that are versatile for various evaluation criteria and align closely with human judgment?}

Designing metrics that meet these criteria involves three essential capabilities: generating input text prompts, understanding human intention, and reasoning about the three-dimensional physical world. 
Fortunately, Large Multimodal Models (LMMs), particularly GPT-4Vision (GPT-4V)~\cite{openai2023gpt},
have demonstrated considerable promise in fulfilling these requirements~\cite{yang2023dawn}. 
Drawing inspiration from humans' ability to perform 3D reasoning tasks using 2D visual information under language guidance, we posit that GPT-4V is capable of conducting similar 3D model evaluation tasks.

In this paper, we present a proof-of-concept demonstrating the use of GPT-4V to develop a customizable, scalable, and human-aligned evaluation metric for text-to-3D generative tasks. 
Building such an evaluation metric is similar to creating an examination, which requires two steps: formulating the questions and evaluating the answers. 
To effectively evaluate text-to-3D models, it is crucial to obtain a set of input prompts that accurately reflect the evaluators' needs. 
Relying on a static, heuristically generated set of prompts is insufficient due to the diverse and evolving nature of evaluator demands. 
Instead, we developed a ``meta-prompt'' system, where GPT-4V generates a tailored set of input prompts based on evaluation focus. 
Following the generation of these input text prompts, our approach involves comparing 3D shapes against user-defined criteria, akin to grading in an exam. 
We accomplish this through designing an instruction template, which can guide GPT-4V to compare two 3D shapes per user-defined criterion. 
With these components, our system can automatically rank a set of text-to-3D models by assigning each of these models an Elo rating.
Finally, we provide preliminary empirical evidence showing that our proposed framework can surpass existing metrics in achieving better alignment with human judgment in a diverse set of evaluation criteria.
Results suggest that our metric can efficiently provide an efficient and holistic evaluation of text-to-3D generative models.

\section{Related Work}
\label{sec:related}
\paragraph{Text-to-3D generation.} 
Text-to-image generation models have become increasingly powerful with text-to-3D extensions being the next frontier (see~\cite{Po:2023:star_diffusion_models} for a recent survey).
However, due to limited amounts of 3D data, text-to-3D has mainly been driven by methods based on optimizing a NeRF representation~\cite{mildenhall2021nerf}.
For example, Dreamfusion~\cite{poole2022dreamfusion} optimizes a NeRF using score-distillation-sampling-based (SDS) loss. 
The quality of such optimization-based methods~\cite{poole2022dreamfusion, sjc, metzer2022latent, Lin_2023_CVPR, chen2023fantasia3d, wang2023prolificdreamer, shi2023mvdream, Tang2023DreamGaussianGG}, however, is far behind that of text-to-image models~\cite{ramesh2021zero, ramesh2022hierarchical, rombach2022high, podell2023sdxl}.
Compared with their 2D counterparts, they are generally lacking diversity, texture fidelity, shape plausibility, robustness, speed, and comprehension of complex prompts. 
On the other hand, Point-E~\cite{nichol2022point} and Shap-E~\cite{jun2023shap} train feed-forward 3D generative models on massive undisclosed 3D data.
Though they show promising results with fast text-to-3D inference, their generated 3D assets look cartoonish without geometric and texture details. 
Recently, we notice a rapid change in the landscape of text-to-3D methods~\cite{liu2023syncdreamer,long2023wonder3d} mainly due to the public release of the large-scale Objaverse datasets~\cite{deitke2023objaverse,deitke2023objaversexl}. 
Feed-forward methods trained on these datasets, e.g., Instant3D~\cite{instant3d2023}, have managed to make a big jump in text-to-3D quality, narrowing the performance gap between 3D and 2D generation. 
As we expect to see continuing progress in this area, it is critical to have robust evaluation metrics closely aligning with human judgment to measure different aspects of 3D generative models, including shape plausibility and texture sharpness.
Such an evaluation metric can provide meaningful guidance for model design choices and support fair comparisons among the research community. 
%
\vspace{-1em}
\paragraph{3D Evaluation Metrics.}
Evaluating 3D generative models is inherently challenging, requiring an understanding of both physical 3D worlds and user intentions. 
Traditional methods for evaluating unconditional or class-conditioned 3D models typically measure the distance between distributions of generated and reference shapes~\cite{Yang2019PointFlow3P,LopezPaz2016RevisitingCT,Achlioptas2017LearningRA,Hao2021GANcraftU3,Bahmani2023CC3DLG,Chan2021EfficientG3}. 
However, these metrics are not readily applicable to text-conditioned generative tasks due to the difficulty in obtaining a comprehensive reference set, given the vastness of natural language inputs ~\cite{bakr2023hrs}. 
To alleviate this issue, prior work tried to curate a set of text prompts to evaluate key aspects of text-conditioned generative tasks~\cite{poole2022dreamfusion,He2023T3BenchBC}.
Our work complements this effort by creating a text-prompt generator using language instruction.
Additionally, prior studies utilized multimodal embeddings, such as CLIP~\cite{CLIP} and BLIP~\cite{Li2022BLIPBL,Li2023BLIP2BL}, to aid the evaluation. 
For instance, the CLIP Similarity metric~\cite{poole2022dreamfusion,jain2022zero} employs CLIP embeddings to assess text-to-3D alignment. 
However, these metrics are often tailored to measure specific criteria, lacking the flexibility to adapt to different requirements of text-to-3D evaluation.
User preference studies are considered the gold standard for evaluating text-to-3D models, as adopted by many papers~\cite{Lin_2023_CVPR,Raj2023DreamBooth3DST,Hllein2023Text2RoomET,Bahmani2023CC3DLG,Seo2023DITTONeRFDI,Tang2023DreamGaussianGG}. 
While user studies offer versatility and accuracy, they are costly, time-consuming, and difficult to scale. 
Our automatic metrics can serve as an alternative to user preference studies, aligning well with human preferences while offering high customizability.
\vspace{-1em}
\paragraph*{Large multimodality models.}
Following the success of large language models (LLMs)~\cite{Brown2020LanguageMA,openai2023gpt,Chowdhery2022PaLMSL,Anil2023PaLM2T,Hoffmann2022TrainingCL,Touvron2023LLaMAOA}, the focus has shifted to large multimodal models (LMMs) as the next frontier in artificial intelligence. 
Initial efforts of LMM involve combining computer vision with LLMs by fine-tuning visual encoders to align with language embeddings~\cite{Tsimpoukelli2021MultimodalFL,Alayrac2022FlamingoAV,Li2023BLIP2BL,Li2022BLIPBL,Huang2023LanguageIN,Driess2023PaLMEAE,Awadalla2023OpenFlamingoAO} or converting visual information to text~\cite{Zeng2022SocraticMC,Wang2022LanguageMW,Hu2022PromptCapPT,Shao2023PromptingLL}.
Most of these models are usually limited in scale.
Recently, GPT-4V~\cite{2023GPT4VisionSC} has risen as the leading LMMs, benefiting from training on an unprecedented scale of data and computational resources. 
These LMMs have demonstrated a range of emerging properties ~\cite{yang2023dawn}, including their capability as evaluators for language and/or vision tasks~\cite{zhang2023gpt,zheng2023judging,hessel2021clipscore}.
In our work, we explore the use of GPT-4V in evaluating 3D generative models, a relatively under-explored application because GPT-4V cannot directly consume 3D information.
%

\section{Method Overview}
\label{sec:method}

The goal of our evaluation metric is to rank a set of text-to-3D models based on user-defined criteria. 
Our method consists of two primary components. 
First, we need to decide which text prompt to use as input for the evaluation task.
Toward this goal, we develop an automatic prompt generator capable of producing text prompts with customizable levels of complexity and creativity (Sec.~\ref{sec:prompts}). 
The second component is a versatile 3D assets comparator (Sec.~\ref{sec:metric}). 
It compares a pair of 3D shapes generated from a given text prompt according to the input evaluation criteria.
Together, these components allow us to use the Elo rating system to assign each of the models a score for ranking (Sec.~\ref{sec:elo}).

\section{Prompt Generation}
\label{sec:prompts}
\begin{figure}
    \centering
    \includegraphics[width=\linewidth]{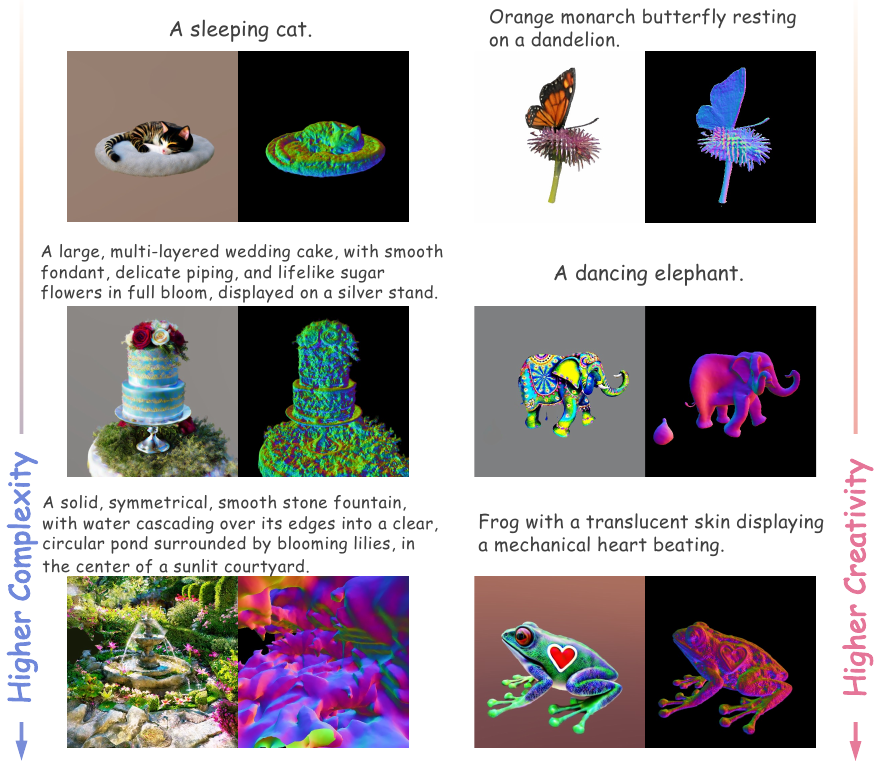}
    \vspace{-2em}
    \caption{
    \textbf{Controllable prompt generator.}
    More complexity or more creative prompts often lead to a more challenging evaluation setting.
    Our prompt generator can produce prompts with various levels of creativity and complexity.
    This allows us to examine text-to-3D models' performance in different cases more efficiently.
}
    \label{fig:prompt_examples}
\end{figure}

Creating evaluation metrics for text-to-3D generative models requires deciding which set of input text prompts we should use as input to these models.
Ideally, we would like to use all possible user input prompts, but this is computationally infeasible.
Alternatively, we would like to build a generator capable of outputting prompts that can mimic the actual distribution of user inputs.
To achieve this, we first outline the important components of an input prompt for text-to-3D models (Sec~\ref{sec:prompt-components}).
Building on these components, we design a ``meta-prompt'' to instruct GPT-4V how to leverage these components to generate an input text prompt for text-to-3D models (Sec~\ref{sec:meta-prompt}).
\begin{figure*}[th!]
    \centering
    \includegraphics[width=\textwidth]{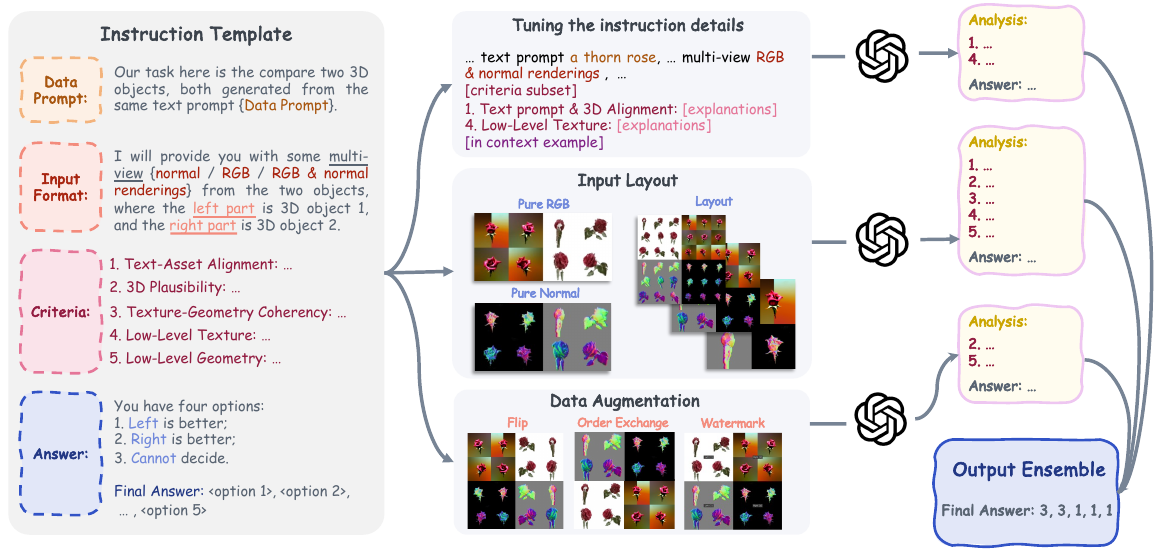}
    \vspace{-1em}
    \caption{
    \textbf{Illustration of how our method compares two 3D assets.} 
    We create a customizable instruction template that contains necessary information for GPT-4V to conduct comparison tasks for two 3D assets (Sec.~\ref{sec:comparator}).
    We complete this template with different evaluation criteria, input 3D images, and random seeds to create the final 3D-aware prompts for GPT-4V.
    GPT-4V will then consume these inputs to output its assessments.
    Finally, we assemble GPT-4V's answers to create a robust final estimate of the task (Sec.~\ref{sec:ensemble})
    }
    \label{fig:pairwise}
    \label{fig:method}
\end{figure*}

\subsection{Prompt components}\label{sec:prompt-components}

A typical input text prompt for text-to-3D models contains three components: subjects, properties, and compositions.
Subjects usually involve nouns referring to objects or concepts the user would like to instantiate in 3D.
``Cats'', ``fire'', and ``universe'' are all examples of subjects.
Properties include adjectives a user can use to describe the subjects or their interactions, such as ``mysterious'' and ``weathered''.
Finally, users will compose these concepts and properties together into a sentence or clause.
The composition varies from as simple as joining different subjects and/or properties together with commas or as thoughtful as writing it as a grammatically correct and fluent sentence.
In this work, we prompt GPT-4V to create a comprehensive list of words for subjects and properties.
This list of subjects and properties will be used as building blocks to construct the ``meta-prompt'', which is an instruction for GPT-4V to generate input text-prompts by composing these building blocks.
Section~\ref{sec:supp-prompt-generator} contains more implementation details.

\subsection{Meta-prompt}\label{sec:meta-prompt}

Provided with ingredients to create input prompts, we now need to automatically compose these ingredients together according to the evaluator-specified requirements.
This requires the prompt generator to understand and follow the evaluator's instruction.
In this paper, we use GPT-4V's ability to generate prompts following instructions.
Specifically, we would like to build a text instruction asking GPT-4V to create a list of prompts that can be used as input for text-to-3D models.
We coin this instruction ``meta-prompt''.

In order for GPT-4V to output prompts for text-to-3D models, we first provide GPT-4V with the necessary ingredients, \ie a list of subjects and properties from the previous section.
In addition to these, the meta-prompt needs to include a description of how the evaluator wants the output prompt set to be.
For example, the evaluator might want to focus on complex prompts containing multiple subject interactions and properties, testing a text-to-3D models' ability to generate complex objects.
One might also be curious about these models' performance in creative prompts involving subjects and descriptions that are not commonly seen in the real world.
How complex and creative the input prompt can influence how challenging the evaluation task is.
These two axes, complexity and creativity, are examples of evaluator's criteria.
Such criteria should be specified as language instructions attached to the ``meta-prompt'' along with all the ingredients.
With both the prompt ingredient and the evaluator's criteria properly included in the meta-prompt, our GPT-4V-based prompt generator can now compose sentences that adhere to the evaluator's requirement.
The appendix contains more details about our meta-prompt and prompt generation pipeline (Sec~\ref{sec:supp-prompt-generator}).

Figure~\ref{fig:prompt_examples} shows prompts outputted from our generator with instruction asking for different complexity and creativity.
We can see that high complexity introduces a larger number of objects, multifaceted descriptions, and occasionally, a completely broken scene. 
Similarly, more creative prompts combine subjects, verbs, or adjectives in unconventional ways.
Text-to-3D models also tend to struggle with these more creative prompts, failing to follow the description of these input prompts exactly.
This suggests that input prompts distribution can greatly affect how challenging the evaluation task is. 
Being able to control the distributions of the input prompt allows us to examine the performance of these text-to-3D models through a more focused lens.

\section{3D Assets Evaluator}
\label{sec:metric}

Now we can sample a set of text prompts, $\tT = \{t_i\}_i$, using our generator.
In this section, we will evaluate the performance of a set of text-to-3D generative models using $\tT$ as input prompts.
Given a set of these models, $\mM=\{M_j\}_j$, we use each model to generate one or more 3D shapes for each prompt. 
This results in a set of tuples: $\{(M_k, t_k, M_j(t_k, \zz_k)) | M_k\in \mM, t_k \in \tT\}_k$, where $\zz_k$ represents the random noise influencing the shape generation. 
Our objective is to rank the text-to-3D models in $\mM$ based on a user-defined criterion. 
To accomplish this, we first prompt GPT-4V to compare two 3D assets generated from the same input text prompt (Sec~\ref{sec:comparator} and Sec~\ref{sec:ensemble}).
We then use these pairwise comparison results to assign each of the models an Elo rating reflecting its performance (Sec~\ref{sec:elo}).

\subsection{Pairwise Comparison }\label{sec:comparator}

At the core of our evaluation metric is the ability to answer the following question: \textit{given a text prompt $t$, and two 3D shapes generated from two different models, say $M_i$ and $M_j$, which 3D shape is better according to the evaluation criteria?}
As discussed in previous sections, we hypothesize that one can leverage GPT-4V to achieve this task. 
However, since GPT-4V is trained on language and visual data, it lacks the ability to analyze 3D shapes directly.
Therefore, our input to GPT-4V should include both text instructions and 2D visual renderings that can capture 3D information.

Specifically, for each of the two 3D assets, we will create a large image containing renderings of the 3D asset from four or nine viewpoints.
These two images will be concatenated together before passing into GPT-4V along with the text instructions.
GPT-4V will return a decision of which of the two 3D assets is better according to the instruction.

\vspace{-1em}\paragraph*{Text instruction.}
We need to communicate three pieces of information for GPT-4V to compare two 3D assets: instructions to complete a 3D comparison task, the evaluation criteria, and descriptions of the output format.
We found it important to emphasize that the provided images are renders from different viewpoints of a 3D object.
In addition to a plain description of the user-defined evaluation criteria, providing instruction about what kind of image features one should use when analyzing for a particular criteria is also useful.
Finally, instead of requesting only the answer of which shape is better directly, we also prompt GPT-4V to explain how it arrives at its conclusion \cite{Wei2022ChainOT, Besta2023GraphOT}.
\vspace{-1em}\paragraph*{Visual features of 3D shapes.}
Once GPT-4V has been prompted to understand the evaluation criteria and task of interest, we now need to feed the 3D shape into the GPT-4V model.
Specifically, we need to create images that can convey the appearance and the geometry features of the 3D shapes.
To achieve that, for each 3D object, we create image renders of the object from various viewpoints.
For each of these viewpoints, we also render a surface normal image.
These normal surface renders will be arranged in the same layout as the RGB render before being fed into GPT-4V.
Using world-space surface normal renders leads to better results because they provide geometric information about the surface and allow reasoning for correspondence between views.
Appendix~\ref{sec:supp-3d-asset-eval} has more implementation details.


\subsection{Robust Ensemble}\label{sec:ensemble}
Even though GPT-4V is able to provide an answer to the pairwise shape comparison problem, its response to the same input can vary from time to time due to the probabilistic nature of its inference algorithm.
In other words, we can view our GPT-4V 3D shape comparator's outputs as a categorical distribution, and each response is a sample from the distribution.
As a result, a single response from GPT-4V might not capture its true prior knowledge since it can be affected by the variance during sampling.
This is particularly the case when the variance of the output distribution is high (\eg, when both choices are equally likely).
Note that this is not a weakness of GPT-4V as similar situations can happen to human annotators when two objects are equally good according to a criterion.
In other words, we are not interested in sampling one instance of how GPT-4V would make a decision.
Instead, estimating with what probability GPT-4V will choose this answer is more useful.

One way to estimate such probability robustly from samples with variance is through ensembling, a technique that has also been explored in other tasks~\cite {yang2023dawn}.
Specifically, we propose to ensemble outputs from multiple slightly perturbed inputs.
The key is to perturb input prompts to GPT-4V without changing the task or evaluation criteria.
The input includes the text instruction, visual images, as well as the random seed.
Our methods deploy different perturbations, including changing random seeds, the layout of renders, the number of rendered views, and the number of evaluation criteria.
Figure~\ref{fig:method} illustrates how we perturb the input and ensemble the results from these perturbed inputs together.  
Appendix~\ref{sec:ablation} includes more details.
\begin{table*}[t]
  \centering
  \small
  \caption{\textbf{Alignment with human judgment (higher is better).} 
  Here we present Kendall's tau ranking correlation~\cite{Kendall1938ANM} between rankings provided by a metrics and those provided by human experts.
  Higher correlation indicates better alignment with human judgment.
  We \textbf{bold-face} the most aligned method and \underline{underline} the second place for each criterion.
  Our method achieves top-two performances for all evaluation criteria, while prior metrics usually only do well for at most two criteria.
  }
  \vspace{-0.5em}
    \begin{tabular}{lccccccc}
    \toprule
    Methods & Alignment & Plausibility & T-G Coherency & Tex Details & Geo Details & Average \\
    \midrule
    PickScore~\cite{kirstain2023pick} & 0.667 & \underline{0.484} & 0.458 & 0.510 & 0.588 & 0.562 \\
    CLIP-S~\cite{hessel2021clipscore} & 0.718 & 0.282 & 0.487 & 0.641 & 0.667 & 0.568 \\
    CLIP-E~\cite{hessel2021clipscore} & \underline{0.813} & 0.426 & \textbf{0.581} & 0.529 & 0.658 &  0.628 \\
    Aesthetic-S~\cite{Schuhmann2022LAION5BAO} &0.795 & 0.410 & \underline{0.564} & 0.769 & \underline{0.744} &  \underline{0.671} \\
    Aesthetic-E~\cite{Schuhmann2022LAION5BAO} & 0.684 & 0.297 & 0.555 & \underline{0.813} & 0.684 &  0.611 \\
    \midrule
    Ours & \textbf{0.821} & \textbf{0.641} & \underline{0.564} & \textbf{0.821} & \textbf{0.795} &  \textbf{0.710}\\
    \bottomrule
    \end{tabular}%
  \label{tab:elo_tau}%
\end{table*}%


\begin{table}[t]
  \centering
  \small
  \caption{\textbf{Pairwise rating agreements (higher is better).}
  We measure the average probability that the decision of the metric matches that of human's for each comparison.
  Our method achieves strong alignment across most criteria. 
  }
    \vspace{-0.5em}
    \begin{tabular}{lrrrrrr}
    \toprule
    Metrics & \multicolumn{1}{c}{Align.} & \multicolumn{1}{c}{Plaus.} & \multicolumn{1}{c}{T-G.} & \multicolumn{1}{c}{Text.} & \multicolumn{1}{c}{Geo.} & \multicolumn{1}{c}{Avg.} \\
    \midrule
    PickS. & \multicolumn{1}{c}{0.735} & \multicolumn{1}{c}{\underline{0.721}} & \multicolumn{1}{c}{0.713} & \multicolumn{1}{c}{0.690} & \multicolumn{1}{c}{\underline{0.740}} & \multicolumn{1}{c}{0.720} \\
    CLIP  & \multicolumn{1}{c}{0.726} & \multicolumn{1}{c}{0.644} & \multicolumn{1}{c}{0.678} & \multicolumn{1}{c}{0.703} & \multicolumn{1}{c}{0.715} & \multicolumn{1}{c}{0.693} \\
    Aest. & \multicolumn{1}{c}{\underline{0.798}} & \multicolumn{1}{c}{0.698} & \multicolumn{1}{c}{\textbf{0.753}} & \multicolumn{1}{c}{\underline{0.817}} & \multicolumn{1}{c}{\textbf{0.780}} & \multicolumn{1}{c}{\underline{0.769}} \\
    \midrule
    Ours & \multicolumn{1}{c}{\textbf{0.810}} & \multicolumn{1}{c}{\textbf{0.826}} & \multicolumn{1}{c}{\underline{0.729}} & \multicolumn{1}{c}{\textbf{0.843}} & \multicolumn{1}{c}{{0.735}} & \multicolumn{1}{c}{\textbf{0.789}} \\
    \bottomrule
    \end{tabular}%
  \label{tab:acc}%
\end{table}%


\subsection{Quantifying Performance}\label{sec:elo}
We have now obtained a list of comparisons among a set of models $\mM$.
The comparisons are over a variety of sampled prompts denoted as $\tT$ according to the user-defined criteria.
Our goal is now to use this information to assign a number for each model in $\mM$ such that it best explains the observed result.
Our quantification method should consider the fact that the comparison results are samples from a probability distribution, as discussed in the previous subsection.
%

This problem is commonly studied in rating chess players, where a game between two players can have different outcomes even if one player is better than the other.
In chess and many other competitions, the Elo score~\cite{elo1967proposed} is perhaps the most widely adapted method to produce a numerical estimation that reflects players' performance.
The Elo rating system has also been adapted in prior works to evaluate image generative models~\cite{shi2020improving, nichol2021glide}.
In this paper, we adapt the version proposed by \citet{nichol2021glide}.
Specifically, let $\sigma_i \in \Re$ denote the Elo score of the $i^{\text{th}}$ model in $\mM$.
A higher score $\sigma_i$ indicates better performance.
We assume that the probability of model $i$ beats model $j$ is:
\begin{align}
    \operatorname{Pr}(\text{``}i \text{ beats } j\text{''}) = \paren{1 + 10^{(\sigma_j - \sigma_i)/400}}^{-1}.
\end{align}
Our goal is to find score $\sigma_i$ that can best explain the observed comparison results given the abovementioned assumption.
This can be achieved via maximum likelihood estimation.
Specifically, let $A$ be a matrix where $A_{ij}$ denotes the number of times model $i$ beats model $j$ in the list of comparisons.
The final Elo score for this set of models can be obtained by optimizing the following objective:
\begin{align}
    \sigma = \argmin_{\sigma} \sum_{i\neq j} A_{ij}\log\paren{1 + 10^{(\sigma_j -\sigma_i)/400}}.
\end{align}
In this paper, we initialize $\sigma_i=1000$ and then use the Adam optimizer~\cite{Kingma2014AdamAM} to minimize the loss to obtain the final Elo score.
Please refer to Sec~\ref{sec:supp-elo} for more mathematical intuition about the formulation of the Elo score.


\section{Results}
\label{sec:t23d-metric}

In this section, we provide a preliminary evaluation of our metric's alignment with human judgment across different criteria.
We first introduce the experiment setup.
We will discuss the main alignment results in Sec.~\ref{sec:alignment-res}.
We then explore how to use our metric to evaluate different models holistically in Section~\ref{sec:holistic-eval}.
Finally, we briefly showcase how to extend our models to different criteria in Section~\ref{sec:additional-criteria}.
\vspace{-1em}
\paragraph*{Text-to-3D generative models to benchmark.}
We involve 13 generative models in the benchmark, including ten optimization-based methods and three recently proposed feed-forward methods.
Please refer to Sec~\ref{sec:more-expr-detail} for the complete list.
We leverage each method's official implementations when available.
Alternatively, we turn to Threestudio's implementation~\cite{threestudio2023}.
For methods designed mainly for image-to-3D, we utilize Stable Diffusion XL~\cite{Podell2023SDXLIL} to generate images conditioned on text as input to these models. 
All experiments are conducted with default hyper-parameters provided by the code.

\vspace{-1em}
\paragraph*{Baselines metrics.}
We select three evaluation metrics with various considerations.
\textbf{1) CLIP similarity} measures the cosine distance between the CLIP features~\cite{CLIP} of the multi-view renderings and the text prompt.
This metric is chosen because it is widely used in previous works as the metric for text--asset alignment~\cite{jain2022zero,poole2022dreamfusion,Hllein2023Text2RoomET}. 
\textbf{2) Aesthetic score}~\cite{Schuhmann2022LAION5BAO} is a linear estimator on top of CLIP that predicts the aesthetic quality of pictures.
We choose this because it is trained on a large-scale dataset.
\textbf{3) PickScore}~\cite{kirstain2023pick} is a CLIP-based scoring function trained on the Pick-a-Pic dataset to predict human preferences over generated images. 
To compute the metrics above, we uniformly sample 30 RGB renderings for each of the generated assets. 
The CLIP similarity and aesthetic score can be directly computed from the multi-view renderings and averaged for each prompt.
Since PickScore takes paired data as input for comparison, we assign 30 paired renderings for each pair of objects before averaging the PickScore results.

\begin{figure*}[t]
    \centering
    \includegraphics[width=0.9\textwidth]{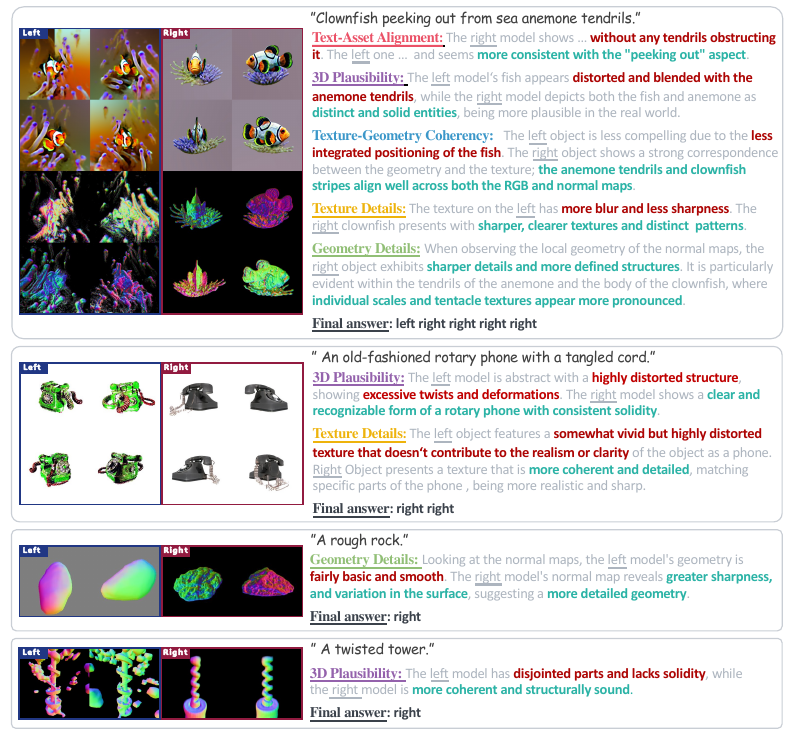}
    \vspace{-1em}
    \caption{\textbf{Examples of the analysis by GPT-4V.} Given two 3D assets, we ask GPT-4V to compare them on various aspects and provide an explanation.
    We find that GPT-4V's preference closely aligns with that of humans. 
    }
    \label{fig:output_examples}
    \vspace{-1em}
\end{figure*}

\vspace{-1em}
\paragraph*{Evaluation criteria.}
%
While our method can potentially be applied to all user-defined criteria, in this work we focus on the following five criteria, which we believe are important for current text-to-3D evaluation tasks.
%
\textbf{1) Text--asset alignment:} how well a 3D asset mirrors the input text description. 
\textbf{2) 3D plausibility:} whether the 3D asset is plausible in a real or virtual environment. 
A plausible 3D asset should not contain improbable parts such as multiple distorted faces (Janus problem) or noisy geometry floaters.
\textbf{3) Texture details:} whether the textures and appearance of the shape are realistic, high resolution, and have appropriate saturation levels. 
\textbf{4) Geometry details:} whether the geometry makes sense and contains appropriate details.
%
\textbf{5) Texture--geometry coherency:} whether geometry and textures agree with each other. For example, eyes of a character should be on reasonable parts of the face geometry.

\vspace{-1em}
\paragraph*{Expert annotation.}
To evaluate the performance of our method, we need to conduct user preference studies to obtain ground truth preference data.
Our user studies will present the input text prompt alongside a pair of 3D assets generated by different methods for the same input.
The user will be asked to identify which 3D asset satisfies the criteria of interest better.
We recruited 20 human experts who are graduate students experienced in computer vision and graphics research to annotate the data. We assigned 3 annotators per comparison question per evaluation criteria.
We compute a reference Elo rating using the formula in Sec.~\ref{sec:elo} using all expert annotations. 


\begin{figure*}
    \centering
    \includegraphics[width=\linewidth]{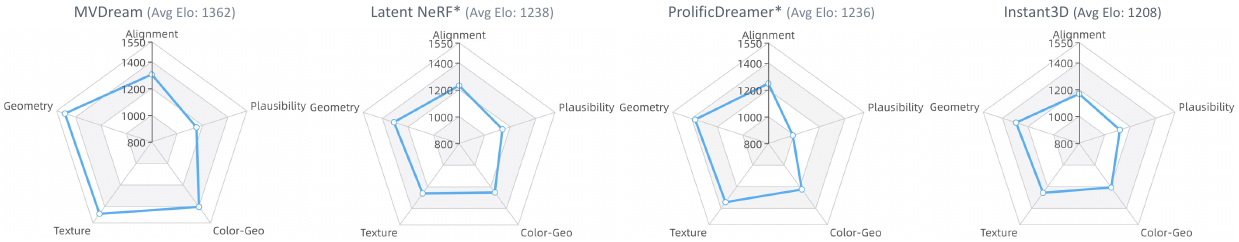}
    \caption{
    \textbf{Holistic evaluation.}
    Since our evaluation metric is human-aligned in multiple criteria, we can evaluate text-to-3D models more holistically. 
    In this figure, we listed the Radar charts of the top four text-to-3D models according to their averaged Elo scores across all five criteria we evaluated. 
    The Radar charts report the Elo rating for each of the five criteria.
    These radar charts can provide relative strengths and weaknesses among these models, providing guidance to improve these models. * indicates results from Threestudio implementation.
    }
    \label{fig:holistic}
\end{figure*}
\begin{figure}[t]
    \centering
    \includegraphics[width=\linewidth]{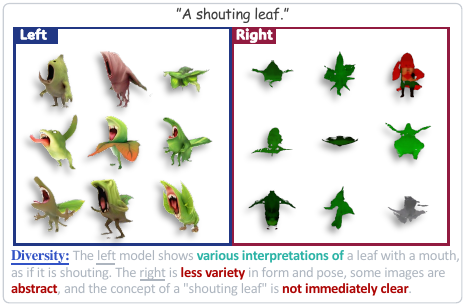}
    \vspace{-2em}
    \caption{
    \textbf{Diversity evaluation.} 
    Our method can be extended to evaluate which text-to-3D models output more diverse 3D assets. 
    }
    \label{fig:diversity}
\end{figure}

\subsection{Alignment with Human Annotators.}\label{sec:alignment-res}
In this section, we evaluate how well our proposed metric aligns with human preference.
To achieve that, we use each metric to assign a score for each text-to-3D model for each evaluation criteria.
Then, we compute Kendell's tau correlation~\cite{Kendall1938ANM} between the scores computed by the metrics and the reference scores.
Table~\ref{tab:elo_tau} shows the ranking correlations between scores predicted by different evaluation metrics and the reference Elo scores computed from expert annotators.
We can see that our metrics achieve the best correlation in 4 out of 5 criteria, as well as the best average correlation.
Note that our method achieves consistent performance across different criteria, while prior metrics usually perform well in only one or two.
This highlights that our method is versatile in different evaluation criteria.

Our metric also shows strong human correlation for each 3D asset comparison question, which is a harder task.
To measure that, we assume the response to each comparison question follows a Bernoulli distribution with probability $p$ to select the first shape.
Let $p_i$ be the probability that the evaluation metric will select the first shape at question $i$ and $q_i$ be that of a human annotation.
We measure the pairwise rating agreement using the probability of a random sample from the metric agreeing with that from a human: $\frac{1}{N}\sum_{i=1}^N p_iq_i + (1-p_i) (1-q_i)$.
Table~\ref{tab:acc} shows that our method achieves top-two agreement across all but one criteria.
Figure~\ref{fig:output_examples} shows some exemplary outputs from our method.
We can see that GPT-4V is also able to provide some analysis justifying its final choice.

\subsection{Holistic Evaluation}\label{sec:holistic-eval}

The versatility of our method lands the provision to paint a holistic picture of each text-to-3D model's performance.
Specifically, we compute each model's average Elo scores across each criterion and present the Radar charts of the models achieving the top averaged Elo scores in Figure~\ref{fig:holistic}. 
According to our metric, MVDream~\cite{shi2023mvdream} won first place on all five criteria.
MVDream achieves about 100 more ELO scores than its runner-ups.
The second, the third, and the fourth places are taken by Latent-NeRF~\cite{metzer2022latent},
ProlificDreamer~\cite{wang2023prolificdreamer}, and Instant3D~\cite{instant3d2023}.
These models achieve similar averaged Elo scores, with differences of less than 30 Elos.
These three models achieve about 100 Elos more than the next tiers of models, which score about 1100 Elos.

While Latent-NeRF, ProlificDreamer, and Instant3D achieve similar overall scores, our metrics allow further analysis into the relative strengths and weakness of each models.
For example, ProlificDreamers show strong performance in three criteria: alignment, geometry details, and texture details.
However, its performance in 3D Plausibility is lagging behind when comparing with the other top-performing models.
Among these three models, Instant3D~\cite{instant3d2023} is a feed-forward method that takes a much shorter time compared to the top two methods.
While our evaluation metrics' reliability can still be limited, we hope that such a holistic picture can provide essential guidance for developing future text-to-3D algorithms.



%
\subsection{Extension to Other Criteria}\label{sec:additional-criteria}
While we focus our empirical studies in five criteria, our metric can be adapted to evaluating a different criteria users might care about.
For example, it is important that a generative model can produce different outputs given different random seeds.
This aspect is commonly underlooked by most text-to-3D metrics. 
With small modification of the text and image prompt input into GPT-4V, our method can be applied to evaluate diversity.
Figure~\ref{fig:diversity} shows the visual image we provide GPT-4V when prompting it to answer the question about which model's output has more diversity.
For each method, we produce 9 3D assets using different random seeds.
We render each of these assets from a fixed camera angle to create the input image fed into GPT-4V.
The text in Figure~\ref{fig:diversity} is an excerpt of GPT-4V's answer.
We can see that GPT-4V is able to provide a reasonable judgment about which image contains renders of more diverse 3D assets.
Currently, we are restricted to qualitative studies because most existing text-to-3D models are still compute-intensive.
We believe that large-scale quantitative study is soon possible with more compute-efficient text-to-3D models, such as Instant3D, becoming available.
\section{Discussion}

In this paper, we have presented a novel framework leveraging GPT-4V to establish a customizable, scalable, and human-aligned evaluation metric for text-to-3D generative tasks. 
First, we propose a prompt generator that can generate input prompts according to the evaluator's needs.
Second, we prompt GPT-4V with an ensemble of customizable ``3D-aware prompts.''
With these instructions, GPT-4V is able to compare two 3D assets according to an evaluator's need while remaining aligned to human judgment across various criteria. 
With these two components, we are able to rank text-to-3D models using the Elo system.
Experimental results confirm that our approach can outperform existing metrics in various criteria.
\paragraph{Limitations and future work.} 
While promising, our work still faces several unresolved challenges.
First, due to limited resources, our experiment and user studies are done on a relatively small scale. 
It's important to scale up this study to better verify the hypothesis.
Second, GPT-4V's responses are not always true.
For example, GPT-4V sometimes shows hallucinations---a prevalent issue for many large pretrained models~\cite{Zhang2023HowLM}. 
GPT-4V can also process some systematic errors, such as bias toward certain image positions~\cite{zheng2023judging,zhang2023gpt}.
Such biases, if unknown, could induce errors in our evaluation metric.
While our ensembling technique can mitigate these issues, how to solve them efficiently and fundamentally remains an interesting direction.
Third, a good metric should be ``un-gamable''.
However one could potentially construct adversarial patterns to attack GPT-4V.
This way one might gain a high score without needing to produce high-quality 3D assets. 
Last, while our method is more scalable than conducting user preference studies, we can be limited by computation, such as GPT-4V API access limits.
Our method also requires a quadratically growing number of comparisons, which might not scale well when evaluating a large number of models when compute is limited.
It would be interesting to leverage GPT-4V to intelligently select input prompts to improve efficiency.

\paragraph{Acknowledgement.} 
This project was in part supported by Google, Samsung, Stanford HAI, Vannevar Bush Faculty Fellowship, ARL grant W911NF-21-2-0104, and Shanghai AI Lab. 
We would love to thank members of Stanford Computational Imaging Lab, Stanford Geometric Computation Group, Shanghai AI Lab, and Adobe Research for useful feedback and discussion. 

{
    \small
    \bibliographystyle{ieeenat_fullname}
    \bibliography{main}
}
\appendix
\setcounter{table}{0}
\setcounter{figure}{0}
\renewcommand{\thetable}{R\arabic{table}}
\renewcommand\thefigure{S\arabic{figure}}

\section{Overview}
This supplementary material includes additional details about our experiment and methods, which can not be fully covered in the main paper due to limited space.
We first provide more details about the method, including our meta-prompts and comparison prompts, in Section~\ref{sec:more-method-detail}.
Experiment details, such as baselines and data, are included in Section~\ref{sec:more-expr-detail}.
We also provide detailed ablation studies about the effectiveness of different ways to prompt GPT-4V and ensemble its output (Section~\ref{sec:ablation}).
More experimental results are provided in Section~\ref{sec:more-results}.
Finally, we demonstrate some failure cases of our methods in Section~\ref{sec:failure-cases}.

\section{Method Details}\label{sec:more-method-detail}
We will include detailed descriptions about how we implement the two components of our method: prompt generator (in Sec.~\ref{sec:supp-prompt-generator}) and 3D assets evaluator (in Sec~\ref{sec:supp-3d-asset-eval}).
Section~\ref{sec:supp-elo} provides additional details about how we use the elo rating.

\subsection{Prompt Generator}\label{sec:supp-prompt-generator}
Our prompt generation pipeline includes a conversation with GPT-4V. 
We aim to design a conversation pipeline that can provide GPT-4V with the necessary context of the prompt generation task while remaining customizable so that evaluators can adapt this conversation to their need.

\begin{figure*}[t]
    \centering
    \includegraphics[width=\linewidth]{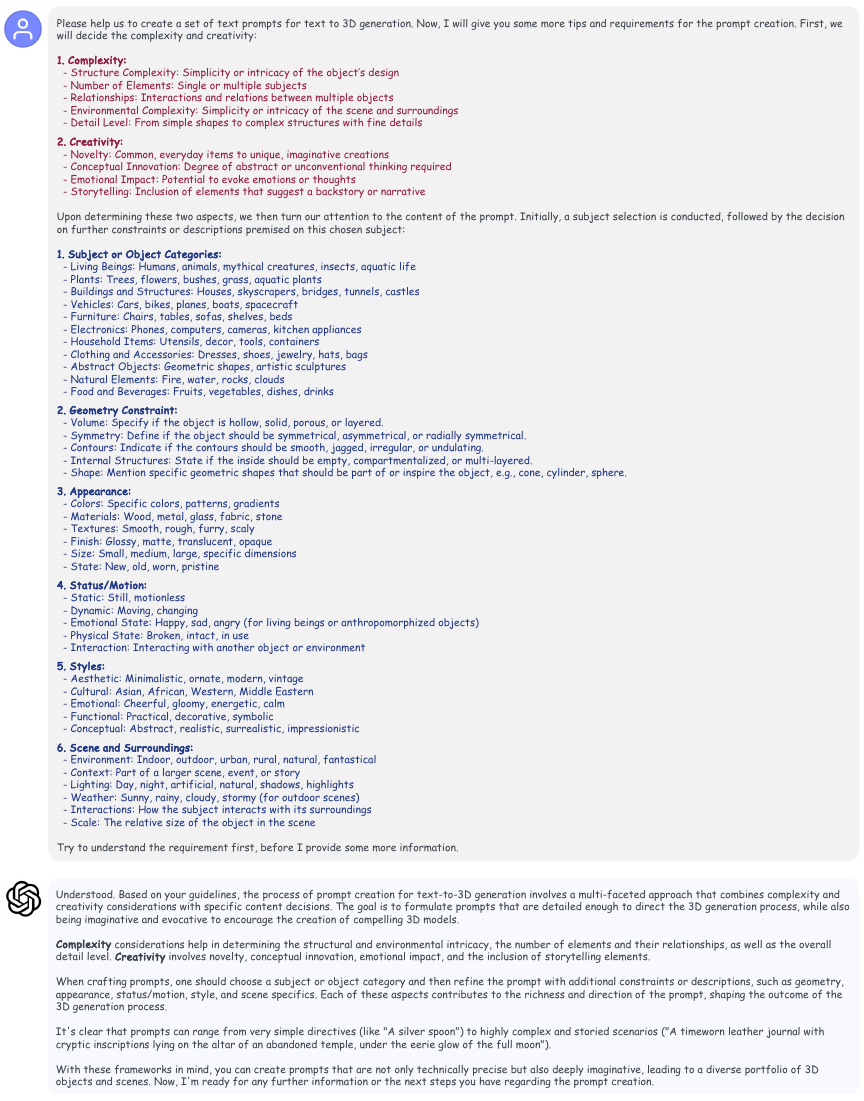}
    \caption{\textbf{Examples of guiding GPT for automatic prompt generation. (Part 1/2)}
    }
    \label{fig:generator_conversion_part_1}
\end{figure*}

\begin{figure*}[t]
    \centering
    \includegraphics[width=\linewidth]{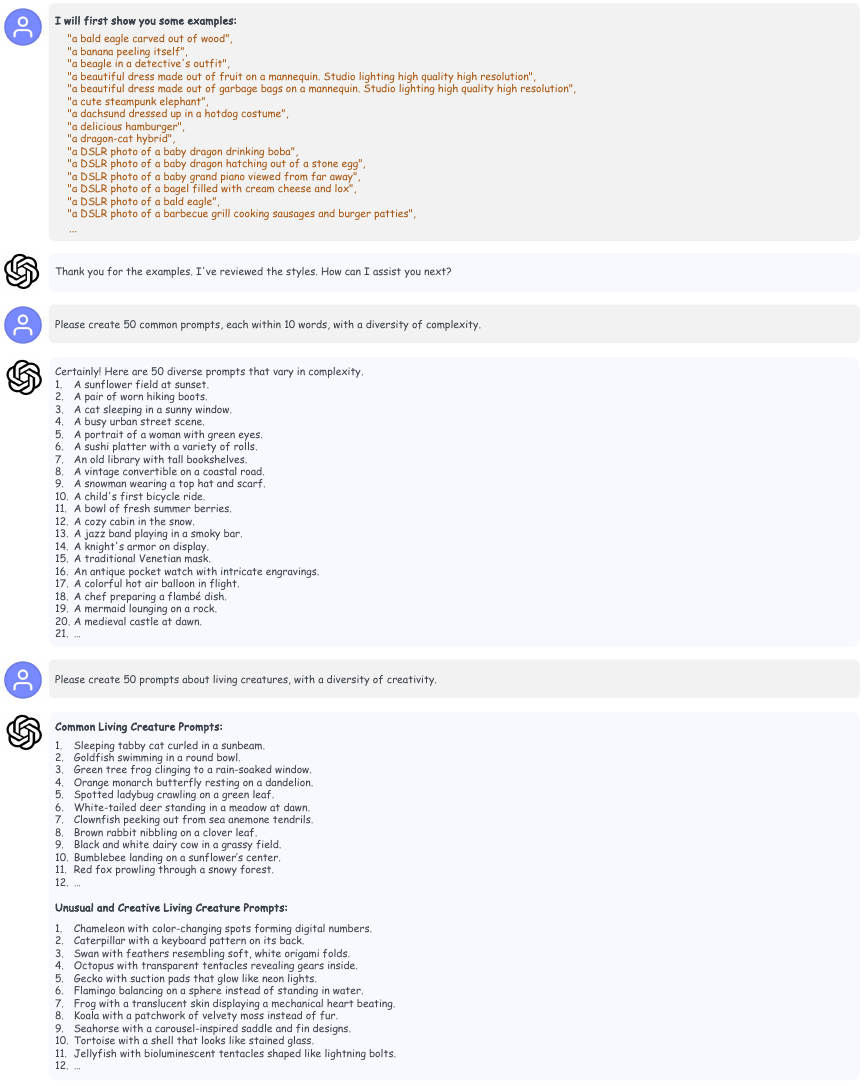}
    \caption{\textbf{Examples of guiding GPT for automatic prompt generation. (Part 2/2)}
    }
    \label{fig:generator_conversion_part_2}
\end{figure*}

Our first prompt describes the task of generating text prompt for text-to-3D generative models.
For example, what's the typical length of each prompt and how do we want the distribution of the final collection of prompts look like.
We also include a description of how an evaluator might want to control this generator in this prompt.
Please see Figure~\ref{fig:generator_conversion_part_1}-\ref{fig:generator_conversion_part_2} for this opening prompt.
\begin{figure*}[t]
    \centering
    \includegraphics[width=\linewidth]{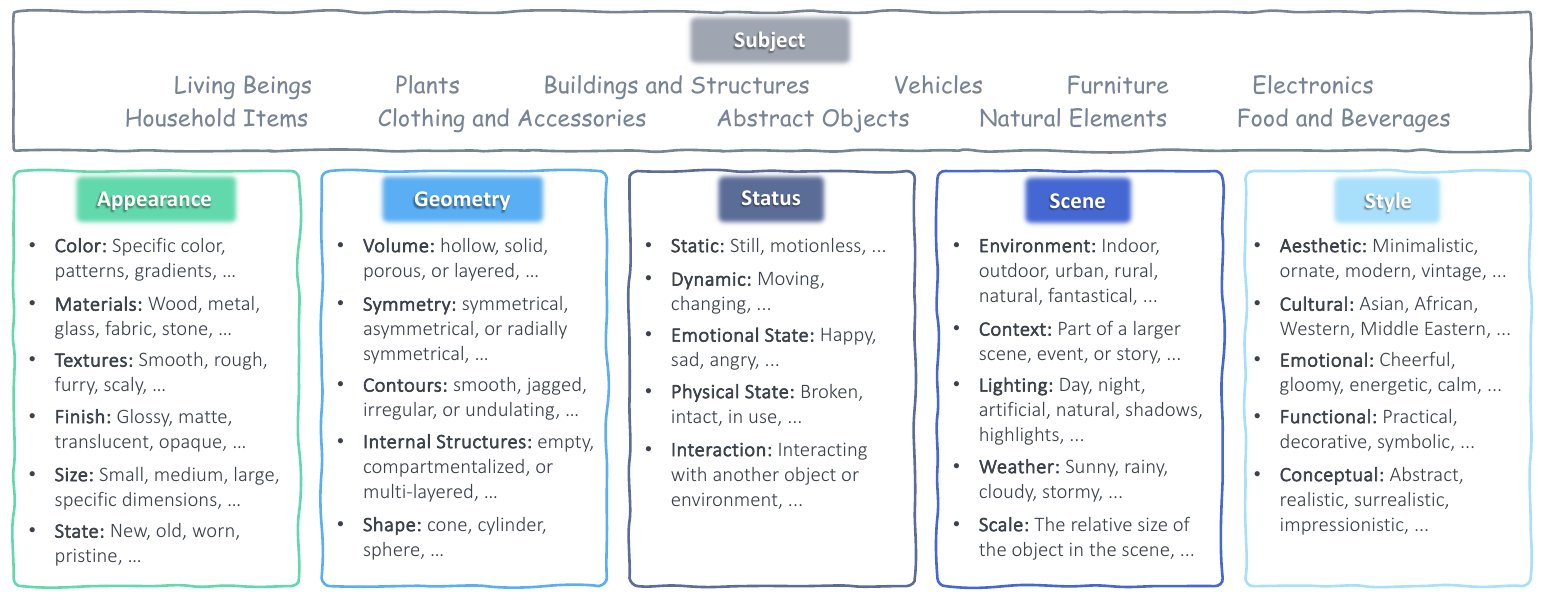}
    \caption{\textbf{Subjects and properties provided for the prompt generator.}
    }
    \label{fig:prompt_for_generator}
\end{figure*}

Now that GPT-4V has some basic understanding of the task, we will further provide it with necessary components needed to composed the prompt.
First, we provide with GPT-4V a list of ten main categories from which it's able to select subjects of the prompt from  (see Figure~\ref{fig:prompt_for_generator}, first box).
These categories are chosen to include the most common classes of objects users are interested in generating in text-to-3D tasks.
The goal is to make our generator aware of a comprehensive list of potential options, reducing the risk of our prompt generator being unintentionally biased toward certain categories when performing in-context learning with exemplary prompts.
These ten categories focuses on what one might wish to generate (\ie the \textit{subjects} mentioned in Section 4.1 of the main paper).

In addition to choosing what to generate, the text-to-3D generative model user also might want to specify a certain state the subject is in (\eg ``a sleeping cat'' as opposed to ``a cat'').
Such description is referred to as the \textbf{properties} in Section 4.1 of the main paper.
To achieve this, we additionally provide GPT-4V a list of descriptions of different properties a user of text-to-3d model might be interested in including (see Figure~\ref{fig:prompt_for_generator}, second row, where each column include list of properties under one of the five aspects we listed).
Note that this specific instruction will base on the subjects and a given level of creativity and complexity the evaluator specifies, which will be attached to the beginning of prompt.
Please see Figure~\ref{fig:generator_conversion_part_1}-\ref{fig:generator_conversion_part_2} for the complete detailed prompt.

Finally, we will provide our prompt generator with a list of exemplary prompt where it can model the output from.
Users can curate this list of prompt according to their need.
In our case, we include the prompts by Dreamfusion~\cite{poole2022dreamfusion}.

\begin{figure*}[t]
    \centering
    \includegraphics[width=\linewidth]{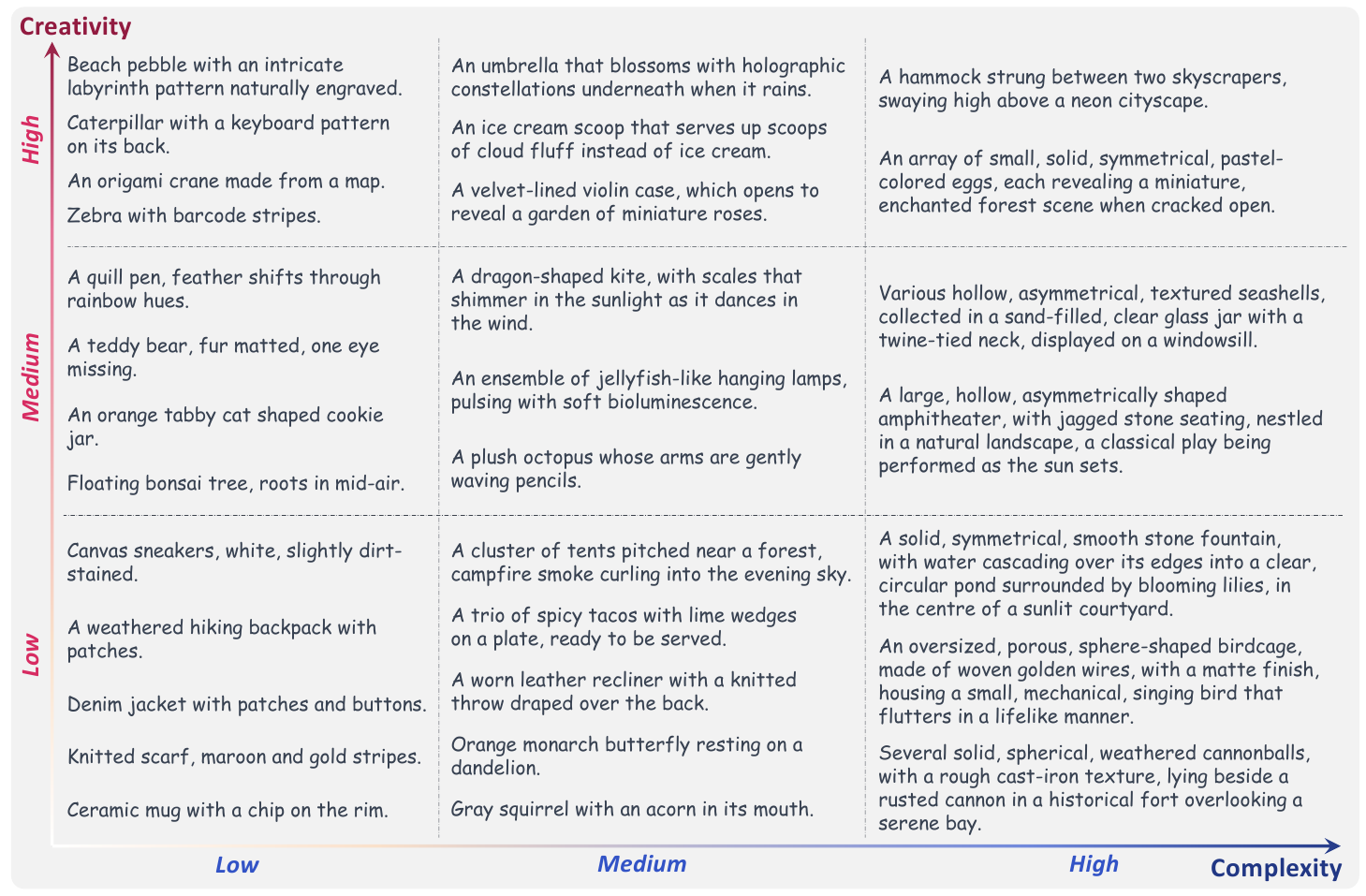}
    \caption{\textbf{Examples of the generated prompts with different levels of creativity and complexity.}
    }
    \label{fig:supp_prompt_examples}
\end{figure*}

With these exemplary prompts and our provided instruction designed to be customizable according to different scenarios, GPT-4V is now able to provide a list of prompts according to the evaluator's input.
Examples of the generated prompts from different levels of creativity and complexity are shown in Figure~\ref{fig:supp_prompt_examples}. 
We can see that users can create prompts with different difficulties and focus using our prompt generator.
While we only focus on only two different axes (\ie creativity and complexity) in this work, but our prompt generating pipeline can be easily adapted to other needs as the evaluators can change various part of our conversations accordingly.


\subsection{3D Assets Evaluator}\label{sec:supp-3d-asset-eval}
%

\begin{figure*}[t]
    \centering
    \includegraphics[width=\linewidth]{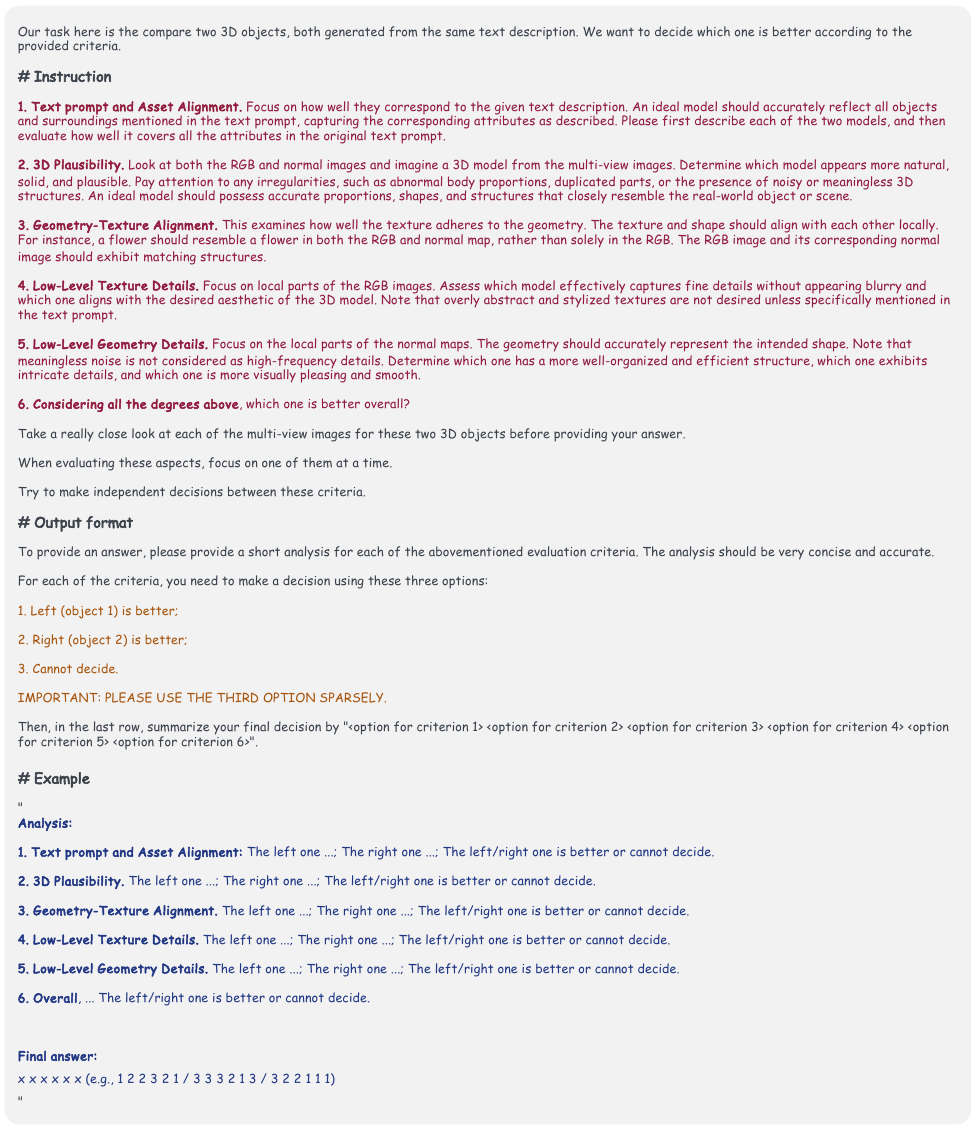}
    \vspace{1em}
    \caption{\textbf{An example of prompts used to guide the GPT-4V evaluator.}
    }
    \label{fig:prompt_for_gpt}
\end{figure*}

%
In the instruction for the GPT-4V evaluator, we first explain the task and the definition of each criterion.
Then, we asked to take a close look at the input pairs of multi-view images and normal maps before providing its analysis and final comparisons. 
Finally, we specify a specific output format for GPT-4V to follow.
In our output format, we also require GPT-4V to provide a short reasoning of why it commit to certain choices, inspired by prompting technique such as chain-of-thought~\cite{Wei2022ChainOT}.
Specifying an output format can also help us better collect results at a larger scale.
An example of such an instruction is shown in Figure~\ref{fig:prompt_for_gpt}.

Note that the description of the criteria can be adjusted according to evaluation task.
In additional, one can also include in-context learning by breaking down this single text-prompt into multiple pieces, where the following pieces provide exemplary answer of the comparison task in hand.
While we believe this can potentially provide a better performance, we only use a single text-prompt and a single image for each pairwise comparison due to limited API access.

In additional to the text instruction, we also need to render images from different view-points to input into GPT-4V.
For each text-to-3D model, we provide one, four, or nine renders with camera evenly distributed surroundingly.
Note that different text-to-3D generative models might use different coordinate systems, and GPT-4V might prefer certain views of the object than others.
We leave it to the evaluator or the text-to-3D model developer to adjust the coordinate system to allow GPT-4V to make the best judgment.

Finally, we create several slightly perturbed version of the input instruction and images to obtain multiple outcomes from GPT-4V.
We will ensemble these outcomes to provide a more accurate estimate, as suggested by~\citet{yang2023dawn}.
Specifically, we perturb the visual information with three versions: including pure RGB renders, pure surface normal maps, and containing both.
For text instructions, we experiment with three versions: 1) GPT-4V is required to evaluate a full list of six criteria, 2) include only the criteria of interest, and 3) explicitly ask GPT-4V to evaluate geometry first.
We also perturb the number of render views in three versions: 1) only a single view-point; 2) four view-points in a 2x2 layout, and 3) 9 view-points in a 3x3 layout.
Finally, we also augment the visual information in three ways: 1) horizontally flipping the left-right order of the objects; 2) vertically flipping the up-down order of the RGB images and normal maps  and 3) adding visual watermark into the image~\cite{yang2023dawn}.
A detailed ablation will be provided in the later section (Sec~\ref{tab:ablation}).

\subsection{ELO Rating}\label{sec:supp-elo}
Comparing two 3D assets can yield different outputs depending on the evaluator or the criteria.
This is because human preference is highly subjective and stochastic.
We need to model such stochasticity when designing a rating that correctly reflects how much a text-to-3D method is preferred by human evaluators.
Such stochasticity is also encountered when two chess players are playing against each other.
Since a stronger player can sill lose to a weaker player when the stronger player he or she is having a bad day.
In other words, the outcome of comparisons can be noisy.
Fortunately, the ELO rating system is designed to create a rating that can reflect the true player's performance while the comparison results are noisy.
We will provide some basic intuition of how our ELO system worked for completeness and please refer to prior literatures for more details~\cite{nichol2021glide,elo1967proposed}.

The ELO rating system tries to assign a number, $\sigma_i$ for the $i^{\text{th}}$ player, which reflects the strength of this player.
Let's examine the case when two players, $i$ and $j$, play together and assume without loss of generality that $\sigma_i > \sigma_j$.
The probability of $i^{\text{th}}$ player winning the game should be larger when the difference between $\sigma_i$ and $\sigma_j$ is larger.
One way to define such probability of winning is the following:
\begin{align}
    \operatorname{Pr}(\text{``} i \text{ beats player } j\text{''}) = \frac{1}{1 + 10^{\frac{\sigma_j - \sigma_i}{c}}},
\end{align}
where $c$ controls the scale of the difference (\ie if the difference is $c$ then the probability of winning will be $1/11$).
From this formula, if $\sigma_i = \sigma_j$, then the probability that player $i$ wins over player $j$ is exactly $1/2$.
It's also easy to see that 
\begin{align}
    &\operatorname{Pr}(\text{``}i\text{ beats } j\text{''}) +  
    \operatorname{Pr}(\text{``}j\text{ beats } i\text{''}) \\
    = &\frac{1}{1 + 10^{\frac{\sigma_j - \sigma_i}{c}}} + \frac{1}{1 + 10^{\frac{\sigma_i - \sigma_j}{c}}} = 1.
\end{align}
During experiment, we observe that $i$ beats $j$ for $A_{ij}$ number of times and $j$ beats $i$ for $A_{ji}$ number of times.
The idea is to find the assignment of $\sigma_i$ and $\sigma_j$ such that it best reflects the empirical observation.
Specifically, we achieve this by maximum likelihood estimation:
\begin{align*}
    &\max_{\sigma_{i,j}} \operatorname{Pr}(A_{ij}, A_{ji} | \sigma_{i}, \sigma_j) \\
    =&\max_{\sigma_{i,j}} \log\paren{
        \operatorname{Pr}(\text{``}i\text{ beats } j\text{''})^{A_{ij}}
        \operatorname{Pr}(\text{``}j\text{ beats } i\text{''})^{A_{ji}}
    } \\
    =& \max_{\sigma_{i,j}} A_{ij}\log\paren{\operatorname{\operatorname{Pr}(\text{``}i\text{ beats } j\text{''})}}
    +
    A_{ji}\log\paren{\operatorname{\operatorname{Pr}(\text{``}j\text{ beats } i\text{''})}} \\
    =& \min_{\sigma_{i,j}} 
    A_{ij}\log\paren{1+10^{\frac{\sigma_j-\sigma_i}{c}}}
    +
    A_{ji}\log\paren{1+10^{\frac{\sigma_i-\sigma_j}{c}}}.
\end{align*}
Equation (1) in the main paper can be directly derived from the equation above by summing over all pairs of $i\neq j$.
In practice, we will initialize $\sigma_i=1000$ for all $i$ and use Adam optimizer to optimize $\sigma_i$ for this loss for $10000$ iterations with a learning rate of $0.1$.
Since Elo score is invariant to adding or subtracting a constant, we further calibrate our scores by setting Dreamfusion~\cite{poole2022dreamfusion} to have an Elo of 1000 for all criteria.

Note that this method consider the setting where the outcomes include only $i$ wins or $j$ wins.
In our case, there are non trivial number of 3D assets pairs from which the human annotator cannot determine which is better.
To handle these cases, we follow \citet{nichol2021glide} to add a win to both methods.
This can effectively dilate the number of winning times.
Our $A_{ij}$ counts the number of times text-to-3D generative model $i$ wins over model $j$ over any captions.
This can be adapted in this theoretical framework by considering the distribution of input text-prompts:
\begin{align*}
    \operatorname{Pr}(\text{``} i \text{ beats player } j\text{''}) =  
    \int \operatorname{Pr}(\text{``} i \text{ beats player } j\text{''}| t) P(t)dt,
\end{align*}
where $t$ denotes an input text prompt.
Following most Elo systems, we choose $c=400$.

\section{Experimental Details}\label{sec:more-expr-detail}
In this section we will provide some additional experiment details.
We provide a detailed list of text-to-3d generative models for benchmarking in Section~\ref{sec:supp-t23d-models}.
Additional detail about our user studies is provided in Section~\ref{sec:supp-user-studies}.

\subsection{Text-to-image Models}\label{sec:supp-t23d-models}
We involve 13 generative models in the benchmark, including ten optimization-based methods and three recently proposed feed-forward methods. 
Please refer to the supplementary for the complete list of methods.
For optimized-base methods, we include DreamFusion~\cite{poole2022dreamfusion}, SJC~\cite{sjc}, Latent-Nerf~\cite{metzer2022latent}, Magic3D~\cite{Lin_2023_CVPR}, Fantasia3D~\cite{chen2023fantasia3d}, Prolific Dreamer~\cite{wang2023prolificdreamer}, DreamGaussian~\cite{Tang2023DreamGaussianGG}, MVDream~\cite{shi2023mvdream}, SyncDreamer~\cite{liu2023syncdreamer} and Wonder3D~\cite{long2023wonder3d}. 
For feed-forward methods, we include PointE~\cite{nichol2022point}, Shap-E~\cite{jun2023shap}, and Instant3D~\cite{instant3d2023}. 
We leverage each method's official implementations when available.
Alternatively, we turn to Threestudio's implementation~\cite{threestudio2023}.
For methods designed mainly for image-to-3D, we utilize Stable Diffusion XL~\cite{Podell2023SDXLIL} to generate images conditioned on text as input to these models. 
Experiments are conducted with default hyper-parameters.


\subsection{User study details}\label{sec:supp-user-studies}

In this paper, we mainly rely on labels provided by expert annotators.
Our expert annotators are all graduate students with computer graphic background (\eg they all have experience looking at surface normal maps).
We recruit twenty such expert annotators from this background.
For 13 methods, we create pairwise comparison between each pair of methods (so $78$ method pairs).
For each method pairs of methods, we sample $3$ captions from the caption generators.
For each of these $234$ comparisons, we assign three different experts to rank all criteria.
The experts are asked to pick a shape that performs better according to certain criteria or indicate that these two shapes are indistinguishable.
Each user will fill out a query form that includes the same description we provided with GPT-4V.
Different from what we provided with GPT-4V, expert annotators are able to see video of 360 rotated render of the object in RGB and in surface normals.
The annotator can control the video (\eg replaying to certain point) to exame the 3D shape in more details.
The video is rendered in 1024x2048 resolution.
In our setting, expert annotators are provided with more information comparing to what we provided to GPT-4V.
Our expert annotators achieve reasonable agreement with greater than 0.53 Cohen kappa~\cite{cohen1960coefficient}.

One disadvantage of expert annotation is that it's difficult to get by so we are only able to obtain a small scale of such annotation.
On the contrary, one can obtain larger-scale annotation from general users.
We've also conducted some preliminary exploration conducting user preference studies with general users.
Unfortunately, we found that data collected from general users are very noisy.
Specifically, we recruited about 53 users and performed the same user studies as done with the expert.
In addition to the instruction provided with GPT-4V, we also provide examples comparison and requires the user to go through a short training session.
The average user agreement (measured by Cohen's kappa) among these users can barely reach 0.3.
This can potentially be caused by the fact that general users do not have experience reasoning about 3D asset information, so computer graphics terms (\eg texture, surface normal, or geometry), become harder to understand.
As a result, we leverage expert user data as reference labels in this paper.
How to conduct large-scale user studies with more available annotators remains a very interesting future work direction.

\section{Ablation Studies}\label{sec:ablation}

In this section, we will examines key technical design choices of our methods.
Specifically, we will focus our ablation on different ways to perturb the input data to create ensemble results (Section 5.2 of main paper).
We first carry out an ablation study on different ways to perturb the input (Section~\ref{sec:supp-inp-perturb}).
Then we show an ablation on how to ensemble these perturbations together (Section~\ref{sec:supp-ensemble}).

Due to limited API access, we are not able to obtain enough GPT-4V queries to compute method-level human alignment scores as in Table 1 or the pair-level alignment score in Table 2 in the main papers.
To address this limitation, we use an alternative metric in the ablation.
For each variant of our methods, we randomly sample 78 different comparisons of 3D assets.
For each comparison, we compute the probability that our GPT-4V 3D asset evaluator would select one shape versus the other.
We denote such empirically estimated probability as $p_i$ for our method's variant to choose one shape at the $i^{\text{th}}$ comparison.
Let $q_i$ be the same probability estimated from expert annotation data.
We return the following L1-distance as an estimation of how our method is \textit{mis}-alignment with human judgment:
\begin{align}
    \operatorname{L1-dist}(p,q) = \frac{2}{N}\sum_{i=1}^N |p_i-q_i|,
\end{align}
where $N$ is the number of comparisons here.
Note that the lower this metric is, the better alignment with human judgement.

\begin{table*}[t]
  \centering
  \small
  \caption{\textbf{Ablation studies on different visual and textual input to GPT-4V.} We mark rank one, rank two, and rank three in each criterion with increasingly lighter shades of blue, and the same baseline (RGB + Normal, 2x2, Joint) is marked in gray.
  }
    \begin{tabular}{c|c|ccccc}
    \toprule
    \multicolumn{2}{c|}{Methods} & Alignment ($\downarrow$) & Plausibility ($\downarrow$) & Color-Geo ($\downarrow$) & Texture ($\downarrow$) & Geometry ($\downarrow$) \\
    \midrule
    \multirow{3}[1]{*}{Visual Information} & Pure RGB & \cellcolor[HTML]{\rankthree}0.523 & \cellcolor[HTML]{\ranktwo}\textbf{0.564} & -     & \cellcolor[HTML]{\rankone}\textbf{0.354} & - \\
          & Pure Normal & 0.674 & 0.654 & -     & -     & 0.579 \\
          & \cellcolor[HTML]{E4E6EA}RGB + Normal & \cellcolor[HTML]{\ranktwo}0.518 & 0.672 & 0.628 & 0.444 & 0.510 \\
    \midrule
    \multirow{3}[1]{*}{Text Instruction} & \cellcolor[HTML]{E4E6EA}Joint & \cellcolor[HTML]{\ranktwo}0.518 & 0.672 & 0.628 & 0.444 & 0.510 \\
          & Separate & \cellcolor[HTML]{\rankone}\textbf{0.451} & 0.597 & \cellcolor[HTML]{\rankthree}0.610 & \cellcolor[HTML]{\rankthree}0.433 & 0.528 \\
          & Geo-first & 0.682 & 0.646 & 0.662 & 0.487 & \cellcolor[HTML]{\rankthree}0.505 \\
    \midrule
    \multirow{3}[4]{*}{View number} & 1     & 0.592 & 0.644 & \cellcolor[HTML]{\ranktwo}0.603 & \cellcolor[HTML]{\ranktwo}0.423 & \cellcolor[HTML]{\rankone}\textbf{0.438} \\
          & \cellcolor[HTML]{E4E6EA}2x2   & \cellcolor[HTML]{\ranktwo}0.518 & 0.672 & 0.628 & 0.444 & 0.510 \\
          & 3x3   & 0.546 & \cellcolor[HTML]{\rankthree}0.582 & 0.654 & 0.503 & 0.559 \\
    \midrule
    \multirow{3}[2]{*}{Augmentation} & Horizontal Flip  & 0.615 & 0.702 & 0.676 & 0.522 & 0.651 \\
          & Vertical Flip & 0.764 & 0.695 & 0.754 & 0.738 & 0.708 \\
          & Watermark & 0.605 & \cellcolor[HTML]{\rankone}\textbf{0.559} & \cellcolor[HTML]{\rankone}\textbf{0.597} & 0.577 & \cellcolor[HTML]{\ranktwo}0.492 \\
    \bottomrule
    \end{tabular}%
  \label{tab:ablation}%
\end{table*}%

\subsection{Ablation for GPT-4V Input Perturbation}\label{sec:supp-inp-perturb}
We conduct ablations on different ways to perturn the inputs for GPT-4V.
We will investigate four different categories of perturbations, including \textit{visual information}, \textit{text instruction}, \textit{view number}, and \textit{augmentation}.
The main results are summarized in Table~\ref{tab:ablation}.

\paragraph{Visual Information}
The visual information can include RGB images, normal maps, or both of them. 
Purely using RGB as input can benefit the perception of texture details, probably because GPT-4V can spend more computational power on the presentation of the textures.
However, not all the criteria can be evaluated merely from the RGB information.
As a result, skip evaluation of alignment on those criteria, namely ``Color-Geo'' and ``Geometry''.
Only presenting the normal maps to GPT-4V does not bring much improvement for its alignment to human choices even for ``Geometry Details''.
We can see that RGB renders seem to play the most important role for GPT-4V to make human-aligned decisions.
Surface normal render is required to perform many evaluations of criteria about geometries.

\paragraph{Text Instruction}
We experiment different way to input user criteria into the text instruction.
Jointly inputting all criteria into a same text prompt can significantly reduce the number of API calls required.
An example of this kind of text instruction can be seen in Figure~\ref{fig:prompt_for_gpt}.
We also try to evaluate only one criterion at a time.
One can see a clear improvement for most of the degrees thanks to the more focused analysis, especially for ``Text-Asset Alignment''.
This presents a trade-off between compute and accuracy.

\paragraph{View number}
The number of views denotes how many multi-view images are shown at the same time.
Given the assumption that the visual context size is fixed for GPT-4V~\cite{2023GPT4VisionSC}, this ablation explores the trade-off between the perception of global coherency and local details.
Presenting only one view at a time can largely improve GPT-4V's ability in evaluating low-level criteria like ``Texture-Geometry Alignment'', ``Texture Details'', and ``Geometry Details''. 
However, the scarcity of views leads to challenges in evaluating the global criteria like ``Text-Asset Alignment'' and ``3D Plausibility''. 
Increasing view numbers to four or nine will largely alleviate this problem.

\paragraph{Augmentation}
In the study, we have also experimented with various visual augmentation techniques~\cite{yang2023dawn}, which refers to changing the visual input slightly without changing the key information contained in the image.
We experiment with three augmentation methods: horizontal flipping of object positions (\ie ``Horizontal Flip''), the rearrangement of RGB images in conjunction with their respective normal maps (\ie ``Verticle Flip''), and the inclusion of watermark annotations to indicate the ``left'' and ``right'' objects (\ie ``Watermark'').
Adding watermarks slightly improves the alignment.
This can be a result of watermarks reducing the ambiguity happened when we refer to certain image positions from the text instruction.

\paragraph{Other findings.}
Another interesting finding is that the results get worse when the normal maps are emphasized in the inputs.
For example, in the setting of ``Geo-first'', we first provide the surface normal and ask questions involving geometries before providing the full RGB renders and asking the rest of questions.
The setting of ``Pure Normal'' also emphasizes surface normal map by not including the RGB renders.
These settings both lead to slightly worse result than the baseline method.


\begin{figure*}[t]
    \centering
    \includegraphics[width=\textwidth]{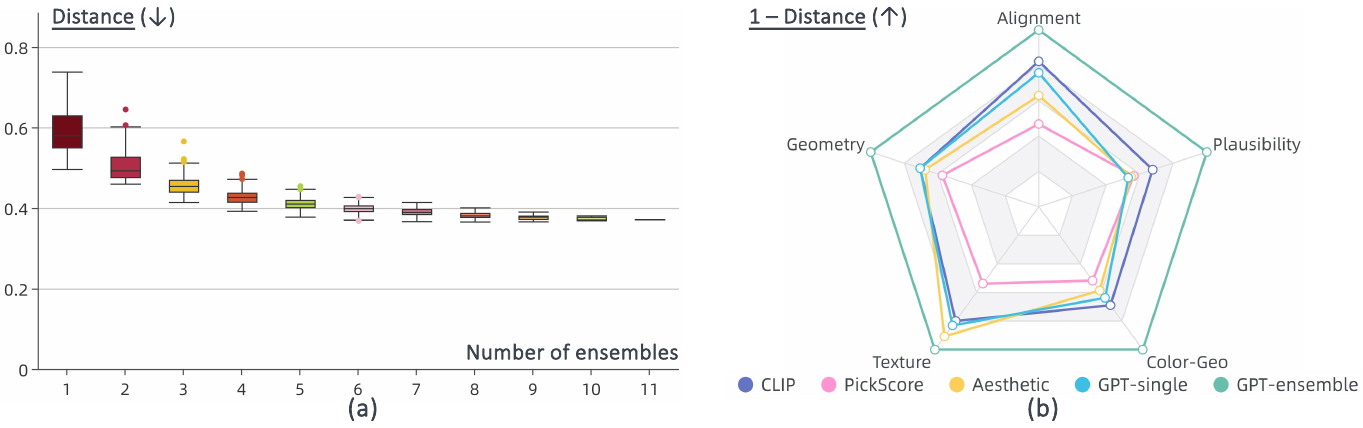}
    \caption{\textbf{Ablation studies of the robust ensemble.} \textbf{(a)} highlights a consistent improvement in performance  with an increase in ensemble size, together with the decrease in the differences among various ensemble strategies. \textbf{(b)} shows how the robust ensemble significantly improves human alignment across all dimensions.
    }
    \label{fig:ablation_ensemble}
\end{figure*}
\subsection{Ablation for Output Ensemble.}~\label{sec:supp-ensemble}
In this section, we want to explore what is the best way to combine different perturbations together.
The exploration space of such combinations is huge as there are $n^k$ numbers of ways to combine $n$ ways to perturb the input together to create $k$ ensembles.
Interestingly, we observed that the variance of the L1-distance reduces as we increase the number of ensembles increases.
This is depicted from Figure~\ref{fig:ablation_ensemble}-(a) shows the distribution of the L1-distance when ensembling different number of results together.
We can see that the alignment performance is not sensitive to particular choice of input perturbations when the number of ensembles is large enough.
To achieve the most effective results, we will incorporate a combination of various view numbers, augmentations, visual information, and individual queries for specific criteria, selecting 4-5 strategies overall for the final combination.
Figure~\ref{fig:ablation_ensemble}-(b) that ensembling these strategies together results in a metric outperforms all the previous metrics when measured in L1-distance.

\section{Additional Results}\label{sec:more-results}
In this section, we will provide additional results which do not fit in the original paper.
\subsection{Ranking}
\begin{table*}[!ht]
  \centering
  \small
  \caption{\textbf{Top-4 methods for different criteria according to our metrics.}}
    \begin{tabular}{cccccc}
    \toprule
          & Alignment & Plausibility & Color-Geo & Texture & Geometry \\
    \midrule
    1st   & MVDream & MVDream & MVDream & MVDream & MVDream \\
    2nd   & Prolific Dreamer & Latent-NeRF & Latent-NeRF & Prolific Dreamer & Prolific Dreamer \\
    3rd   & Latent-NeRF & Instant3D & Prolific Dreamer & Latent-NeRF & Latent-NeRF \\
    4th   & Instant3D & Dreamfusion & Instant3D & Instant3D & Instant3D \\
    \bottomrule
    \end{tabular}%
  \label{tab:ranking}%
\end{table*}%

\begin{figure*}[ht]
    \centering
    \includegraphics[width=0.89\textwidth]{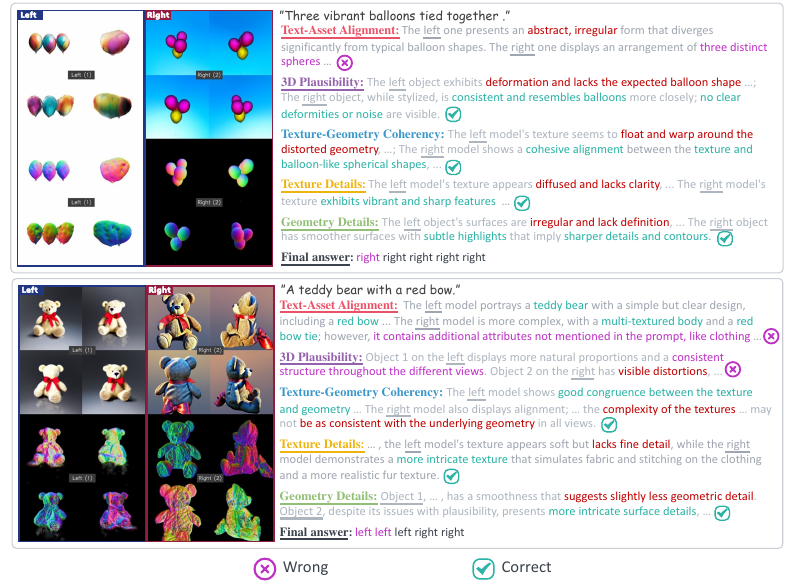}
    \caption{\textbf{Failure cases.} The analysis of GPT-4V can sometimes be \textit{partially} wrong. We show several typical examples for the ``Text-Asset Alignment'' and ``3D Plausibility'' criteria.
    }
    \label{fig:failure_cases}
\end{figure*}
While the results of our method are still largely limited by API request limits, we would like to show some preliminary results for how existing text-to-3D generative models perfrm according to our methods in different criteria.
Table~\ref{tab:ranking} shows the top four methods according to GPT-4V in all six criteria.
We can see that MVDream~\cite{shi2023mvdream} is ranked the first across various criteria.
Latent-NeRF~\cite{metzer2022latent} is able to achieve strong performance, ranked second or third across all criteria.
Similarly, Prolific dreamer~\cite{wang2023prolificdreamer} achieves comparable performance as latent-NeRF except for Plausibility.
Finally, Instant3D~\cite{instant3d2023} is ranked the fourth places in all but Plausibility.
Instant3D is ranked the third place in Plausibility, above Prolific dreamer.
Dreamfusion~\cite{poole2022dreamfusion} is able to achieve good Plausibility, probably thanks to its geometry regularization.

%

%
\subsection{GPT-4V Comparison Examples}
\begin{figure*}[ht]
    \centering
    \includegraphics[width=0.89\textwidth]{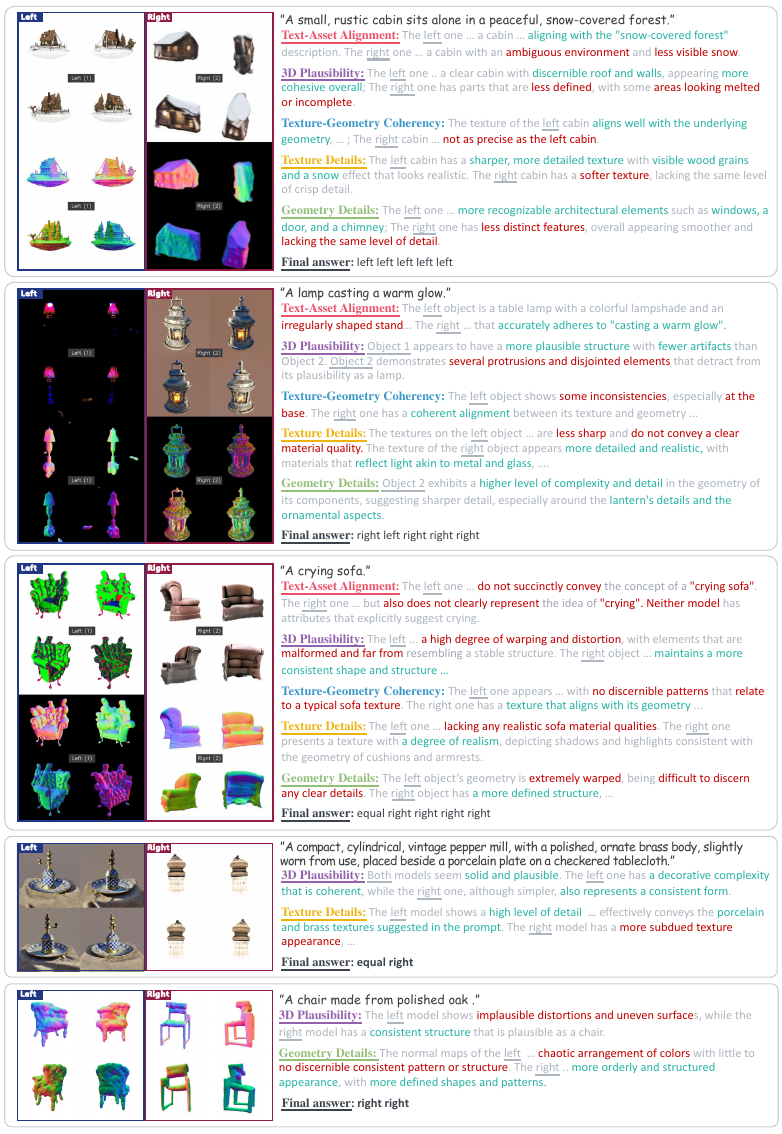}
    \caption{\textbf{Additional examples of the analysis by GPT-4V.} 
    }
    \label{fig:more_output_examples}
\end{figure*}

In Figure~\ref{fig:more_output_examples}, we show some more examples of GPT-4V's analysis and answers to this task (using ``left / right / equal'' instead of ``1 / 2 / 3'' for better clarity. 
These examples demonstrate how the evaluator is able to make comparisons of the paired visual input based on close observations and detailed analysis.

\section{Failure cases}\label{sec:failure-cases}
We present some typical failure cases in Figure~\ref{fig:failure_cases}. 
In the first instance, GPT-4V fails to detect the mismatch in the number of balloons between the two objects, thereby highlighting its limitations in accurately counting specific quantities. 
In the second instance, both objects exhibit significant issues with their underlying geometries,
The object on the left presents severe 3D plausibility problems, as evidenced by the presence of multiple faces and leg.
Both objects are plagued by low-level geometric noise. 
GPT-4V demonstrates sensitivity to low-level noise, consequently overlooking the overarching issue with the object on the left. 
Such problems can potentially be rectified if one can provide a larger number of views to GPT-4V.

\end{document}